# IceWatch: Forecasting Glacial Lake Outburst Floods (GLOFs) using Multimodal Deep Learning

Zuha Fatima, Muhammad Anser Sohaib, Muhammad Talha, Ayesha Kanwal, Sidra Sultana, Nazia Perwaiz

*Abstract*—Glacial Lake Outburst Floods (GLOFs) pose a serious threat in high mountain regions. They are hazardous to communities, infrastructure, and ecosystems further downstream. The classical methods of GLOF detection and prediction have so far mainly relied on hydrological modeling, threshold-based lake monitoring, and manual satellite image analysis. These approaches suffer from several drawbacks: slow updates, reliance on manual labor, and losses in accuracy when clouds interfere and/or lack on-site data. To tackle these challenges, we present IceWatch: a novel deep learning framework for GLOF prediction that incorporates both spatial and temporal perspectives. The vision component, RiskFlow, of IceWatch deals with Sentinel-2 multispectral satellite imagery using a CNN-based classifier and predicts GLOF events based on the spatial patterns of snow, ice, and meltwater. Its tabular counterpart confirms this prediction by considering physical dynamics. TerraFlow models glacier velocity from NASA ITS_LIVE time series while TempFlow forecasts near-surface temperature from MODIS LST records; both are trained on long-term observational archives and integrated via harmonized preprocessing and synchronization to enable multimodal, physics-informed GLOF prediction. Both together provide cross-validation, which will improve the reliability and interpretability of GLOF detection. This system ensures strong predictive performance, rapid data processing for real-time use, and robustness to noise and missing information. IceWatch paves the way for automatic, scalable GLOF warning systems. It also holds potential for integration with diverse sensor inputs and global glacier monitoring activities.

## I. INTRODUCTION

GLACIAL Lake Outburst Floods (GLOFs) are sudden, large releases of water from lakes dammed by glaciers or moraines. These events often occur due to glacier retreat, dam erosion, or accelerated melting [1]. They can discharge millions of cubic meters in just hours, reaching peak flows of up to 15,000 m³/s [2]. This poses serious threats to communities, infrastructure, and ecosystems in mountain regions such as the Himalayas and the Andes. Detecting GLOFs is essential to reduce risk, protect lives, and improve climate resilience in high mountain environments.

Traditional methods for monitoring GLOFs depend on hydrological breach models, threshold-based water-land indices (like NDWI/MNDWI), and expert analysis of satellite images [3]–[5]. Standard machine learning techniques, such as SVM and Random Forest, have detected glacial lakes, especially in the Karakoram [6]. However, they often struggle with small lakes (< 0.01 km²), cloud cover, and varied terrain. Deep segmentation models like U-Net and DeepLab improve spatial accuracy (IoU > 0.85) [7], [8], but they are mainly focused on mapping rather than predicting GLOF risk [9]. Furthermore, these vision-only methods often lack clarity and do not integrate well with physical glacier dynamics [10].

Recent research spotlights the potential of combining multi-sensor data and deep architectures for glacier monitoring. Reviews outline the revolutionary impact of machine learning for mapping glacial lakes with remote sensing during 2010–2025 [11], including both optical (Sentinel-2/Landsat) and SAR sources, despite adverse conditions given by cloud coverage. More recent models include GlacierNet2 (IoU ≈ 0.86), which combines CNN segmentation with hydrological flow post-processing in debris-covered glaciers [12]; deep learning-based velocity estimation through image matching, namely, ICEpy4D, outperforms classical approaches for glacier motion detection [13]; transformer segmentation, reaching generalization IoU > 0.85 for large-scale variations [14]. However, all the above approaches remain purely mapping-centric and do not predict the dynamic GLOF risk.

In that respect, we introduce IceWatch, a predictive and interpretable dual-stream framework for GLOF detection. The RiskFlow vision stream is designed to capture relevant visual cues such as glacial lake formation, fractures, and meltwater patterns by using a Sentinel-2-based CNN. The TerraFlow tabular stream utilizes a Transformer model trained on NASA ITS_LIVE velocity data to predict glacier movement, whereas TempFlow uses a bidirectional LSTM for the prediction of 30-day temperature trends from MODIS LST data, capturing thermal signals indicating lake instability.

IceWatch offers significant improvements over the earlier frameworks by addressing a number of critical limitations. It improves the small-lake and fine-edge detection by leveraging the more refined LinkNet50 with DenseCRF, attaining high performance metrics: ≈96% recall, ≈92% precision, ≈94% F1-score, and ≈90% IoU in lakes as small as 160 m² [8]. Finally, it minimizes cloud-related errors with effective cloud-masking of Sentinel-2 images and temporal temperature modeling that

This work is supported by National University of Sciences and Technology, Islamabad 44000, Pakistan. *(Corresponding author: Nazia Perwaiz)*

Zuha Fatima, Muhammad Anser Sohaib, and Muhammad Talha are with the National University of Sciences and Technology, Islamabad 44000, Pakistan (email: msohaib.bee21seecs@seecs.edu.pk; zfatima.bee21seecs@seecs.edu.pk; talha.bee21seecs@seecs.edu.pk).

Ayesha Kanwal, Sidra Sultana, and Nazia Perwaiz are with the National University of Sciences and Technology, Islamabad 44000, Pakistan (e-mail: ayesha.kanwal@seecs.edu.pk; sidra.sultana@seecs.edu.pk; nazia.perwaiz@seecs.edu.pk ).

allows the production of reliable predictions under occluded conditions [6], [11]. Furthermore, IceWatch introduces dynamic forecasting through TerraFlow and TempFlow, which model glacier velocity and surface temperature trends respectively, adding physical grounding and predictive capability to the system. Its dual-stream architecture enables cross-verification between spatial and temporal predictions, improving interpretability and addressing the black-box nature often associated with deep learning models [10], [13].

While works such as those summarized in major remote sensing reviews [11] and multilayer segmentation studies [12], [14] excel in spatial mapping, they do not couple forecasting with physical validation. Studies on deep CNN prediction of fluvial floods demonstrate speed and accuracy improvement over hydraulic models [15], while unsupervised ML approaches for ice flow prediction [16] highlight the potential for future-frame modeling. IceWatch uniquely fuses spatial CNN detection with Transformer- and LSTM-based tabular models for real-time GLOF detection with physical validation, not achieved in prior literature.

Building upon our earlier work, GLOFNet [35], which introduced a harmonized multimodal dataset integrating Sentinel-2 imagery, NASA ITS_LIVE glacier velocity fields, and MODIS Land Surface Temperature records, IceWatch extends this foundation by developing a deep learning framework for predictive GLOF detection and early warning.

The main contributions of this article are as follows.
1) IceWatch introduces a multimodal architecture combining a CNN-based spatial stream with two time-series forecasting models, TerraFlow (Transformer-based glacier velocity prediction) and TempFlow (LSTM-based temperature forecasting). This design shifts GLOF detection from retrospective classification to anticipatory risk prediction by bridging visual cues with dynamic physical processes, leading to more accurate and interpretable results.
2) Building on GLOFNet's [33] multimodal backbone, IceWatch utilizes harmonized remote sensing datasets (Sentinel-2, ITS_LIVE, MODIS) for preparing the data and performing cross-modal validation. This ensures that the visual detections are strongly supported by independent physical and thermal evidence.
3) The framework represents a novel multimodal, physics-informed approach to GLOF forecasting; rarely explored in prior literature; combining spatial, thermal, and dynamic glacier indicators within an interpretable, cloud-ready design for scalable early-warning deployment.

The rest of this article is organized as follows. Section II reviews the related works. Section III describes the proposed method in detail. To evaluate our method, experiments are designed in Section IV, and the proposed method is discussed. Finally, Section V concludes this article

## II. RELATED WORKS

Glacial Lake Outburst Flood (GLOF) detection has been approached from diverse angles over the past two decades, ranging from empirical hydrological modeling and manual remote sensing interpretation to recent applications of machine learning and deep learning. While these methods have improved lake mapping accuracy and expanded global GLOF inventories, they often lack predictive capability, physical interpretability, or operational scalability. In this section, we review prior research in five key areas relevant to our work: traditional and machine learning-based GLOF mapping, deep learning for glacier and lake segmentation, time-series modeling of glacier dynamics, hybrid risk assessment frameworks, and the remaining gaps that motivate the IceWatch system.

*A. Static GLOF Mapping and Classical ML-Based Detection*

Initial GLOF assessments relied heavily on manual mapping, empirical breach modeling, and static lake inventories [18], [3]. While these efforts, such as those by Mool et al. and Allen et al., formed the basis for regional hazard frameworks, their static nature limits scalability and real-time applicability. Accordingly, machine learning techniques like the Support Vector Machines and Random Forests were proposed to map glacial lakes from Landsat and ASTER imagery [6], but they lacked adaptability across terrain types and temporal robustness. Similarly, OBIA also became popular for the classification of glacier surfaces; however, it was sensitive to noise and required significant manual tuning [7].

*B. Deep Learning for Glacier and Glacial Lake Segmentation*

CNN-based architectures like U-Net, DeepLab, and LinkNet have significantly pushed forward glacier and glacial lake mapping with IoU scores higher than 85% being reported in [7], [8]. GlacierNet2 presented a hybrid CNN framework that integrated post-processing to improve the delineation of debris-covered ice [8]. Also, different variants of the U-Net architecture attained a maximum segmentation accuracy of 97% for the Bara Shigri and Chandra-Bhaga basins [13], [15]. Recent works have explored attention-based and multi-sensor fusion networks incorporating Sentinel-2, Sentinel-1, and DEMs to generalize glacier boundary detection for different geographic regions [30], [33]. Sharma et al. proposed an improved way of finding small lakes with the help of satellite images by including deep learning-based methodologies which were oriented toward spectral anomalies [31]. Despite all these works, the methods mentioned above remained static and focused on segmentation without any temporal and predictive modeling.

*C. Forecasting Glacier Dynamics Using Time-Series Deep Learning*

Deep learning approaches for time-series modeling, especially LSTM networks, have been used to forecast snow cover, temperature, and precipitation, outperforming classical statistical models like ARIMA [24]. TerraFlow builds upon this by introducing Transformer-based velocity forecasting, which has not previously been applied in glaciology. ICEpy4D demonstrated velocity estimation using optical image pairs and CNNs [13], while Varshney et al. used fully convolutional networks to estimate ice thickness profiles [32]. GANs and unsupervised learning have also been explored to simulate glacier flow patterns and surface dynamics [16]. These studies highlight the potential of temporal modeling but do not integrate visual evidence or deliver interpretable, real-time GLOF predictions.

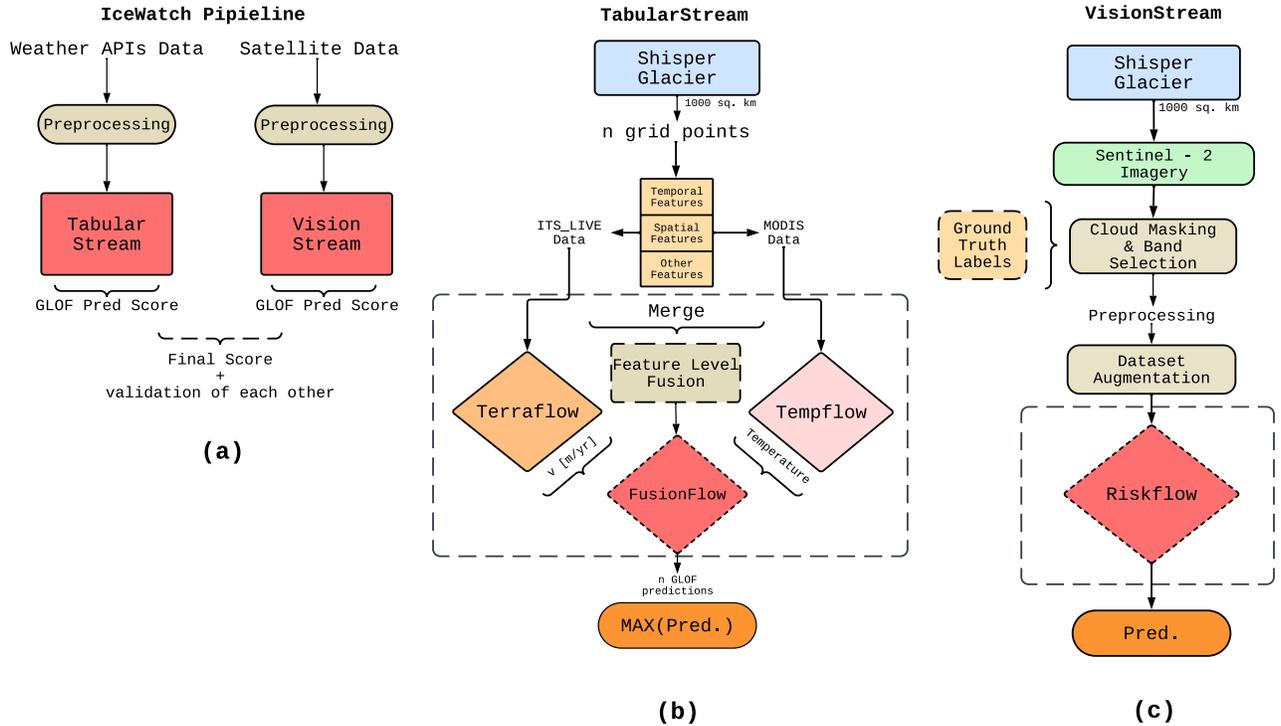

Fig. 1. The IceWatch pipeline for GLOF prediction: (a) dual-stream framework combining tabular and vision data, (b) TabularStream with TerraFlow, TempFlow, and FusionFlow for feature-level fusion, and (c) VisionStream using Sentinel-2 imagery and RiskFlow for final classification.

*D. Hybrid and Multi-Modal Frameworks for GLOF Risk Assessment*

Few studies have attempted to combine spatial and physical indicators for GLOF risk. El Nadi et al. proposed a hybrid framework correlating glacier surge velocity with lake volume to predict debris flows in the Karakoram [26]. Abbas et al. combined machine learning with process-based indicators to predict GLOF-induced debris flows, thereby obtaining improved spatial resolution of the risk. Nurakynov et al. merged InSAR-derived deformation, densityand elevation-based thinning metrics from DEMs, and lake area expansion metrics within a multisensor framework in order to assess the GLOF hazard at Galong Co. Simultaneously, Tom et al. used machine learning on global glacial lake inventories; however, their work focused on the retrospective classification of those features instead of prospectively forecasting. Li et al. provided a comprehensive review of remote-sensing approaches, although mainly focused on mapping rather than prediction. While these recently developed frameworks bring in various types of inputs, they do not include real-time temporal forecasting and do not operationalize real-time AI-driven physical validation.

In this paper, we present a novel dual-stream deep learning framework for the early detection of GLOFs. The system consists of a Vision Stream, where a convolutional neural network called RiskFlow analyzes Sentinel-2 imagery, and a Tabular Stream, where TerraFlow predicts glacier velocity from ITS_LIVE data and TempFlow predicts surface temperature from MODIS observations. By integrating visual and physical signals, IceWatch provides cross-validated, interpretable risk assessments to overcome the single-modality limitations and improve reliability in early warning.

III. METHODOLOGY

The IceWatch framework is designed to combine visual evidence with physical modeling for reliable GLOF prediction. All datasets used in IceWatch modules were preprocessed according to the standardized pipeline described in GLOFNet [35]. Its dual-stream architecture (fig. 1(a)) can be summarized as follows:

1) Vision Stream (fig. 1(c)) comprises of an image-driven module, RiskFlow (fig. 2(c)), that uses Sentinel-2 multispectral imagery and a CNN-based classifier to detect signs of glacial lake formation and potential outburst conditions.
2) The Tabular Stream (fig. 1(b)), models glacier dynamics from geospatial and climatic time series, enabling validation of visual predictions with measurable

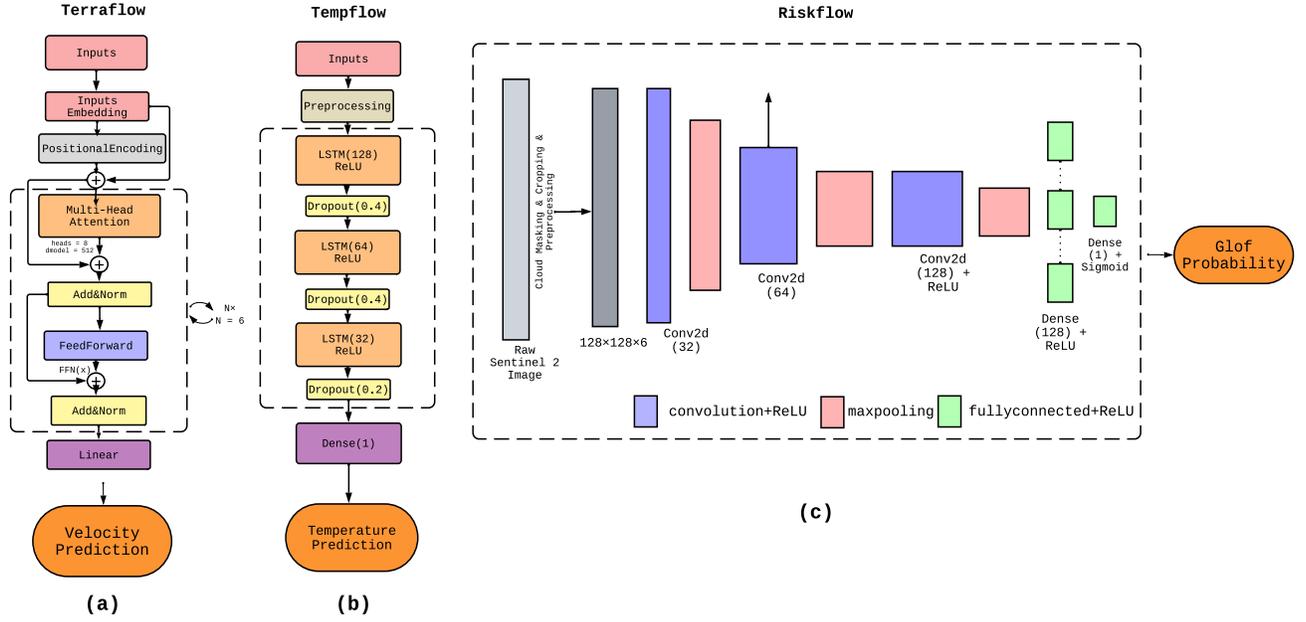

Fig. 2. Multimodal architecture of the proposed IceWatch framework comprising (a) TerraFlow, a Transformer-based model forecasting glacier velocity from NASA ITS_LIVE data; (b) TempFlow, an LSTM-based network predicting near-surface temperature from MODIS LST observations; and (c) RiskFlow, a CNN-based vision model analyzing Sentinel-2 imagery to identify visual indicators of potential GLOFs. Together, these modules fuse spatial and temporal modalities for physics-informed, interpretable GLOF prediction.

glaciological evidence. It consists of two predictive models:

a) TerraFlow (fig. 2(a)) is a transformer-based prediction of glacier velocity using NASA ITS_LIVE velocity products.
b) Tempflow (fig. 2(b)) is an LSTM-based forecasting of near-surface temperature using NASA MODIS LST data.
c) FusionFlow is a feature-level fusion mechanism that integrates predictions from both streams, ensuring that visual anomalies detected by RiskFlow are validated against measurable changes in velocity and temperature.

The IceWatch framework leverages three complementary datasets: Sentinel-2 multispectral imagery, NASA ITS_LIVE glacier velocity products, and MODIS Land Surface Temperature (LST) data to capture visual and physical indicators of GLOF risk in the Shisper Glacier. The data collection, preprocessing, and harmonization protocols follow the procedures established in GLOFNet [35], which provides detailed documentation of cloud masking, normalization, temporal interpolation, augmentation, and multimodal alignment for these datasets. Sentinel-2 multispectral imagery was employed for the Vision Stream due to its 10–20 m spatial resolution and coverage of key cryospheric bands sensitive to ice, snow, vegetation, and meltwater. For the Tabular Stream, NASA's ITS_LIVE velocity product provided multi-decadal glacier surface velocity measurements, while MODIS Land Surface Temperature (LST) data offered daily thermal observations from 2000 to 2024. Preprocessing included cloud masking, normalization, and image resizing for Sentinel-2, followed by augmentation to address class imbalance. The ITS_LIVE dataset underwent aggregation, outlier removal, and cyclical encoding of temporal features to preserve seasonal trends, while MODIS data required quality-flag filtering, interpolation of missing values, and normalization. These steps ensured that all input streams were harmonized, noise-reduced, and suitable for deep learning architectures requiring both temporal continuity and spatial consistency.

*A. RiskFlow (Vision Stream)*

The vision stream called RiskFlow, aims to catageorize Se--ntinel-2 satellite imagery into either GLOF (Glacial Lake Outburst Flood) or non-GLOF classifications. Sentinel-2 was chosen for its high spatial resolution (10–20 m) and extensive multispectral coverage, making it particularly effective for tracking changes in cryospheric areas. Six spectral bands were selected: B2 (Blue), B3 (Green), B4 (Red), B8 (NIR), B11 (SWIR-1), and B12 (SWIR-2). These bands facilitate the differentiation between snow, ice, vegetation, bare soil, and water bodies, which are crucial for understanding glacier-lake dynamics. The preprocessing steps included extracting the bands, normalizing pixel values to the range [0,1], resizing the images to 128×128 pixels, and applying light augmentations such as flips, rotations, and brightness adjustments to reduce overfitting. An area from Shisper Glacier (Fig. 5 and Fig.6) was chosen because of high GLOF occurances in respective area.

One major obstacle was the significant class imbalance.

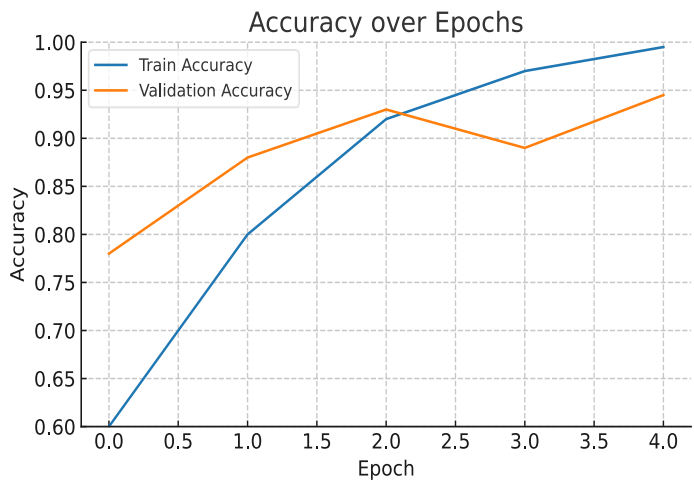

Fig. 3. Training and Validation Accuracy Over Four Epochs for RiskFlow

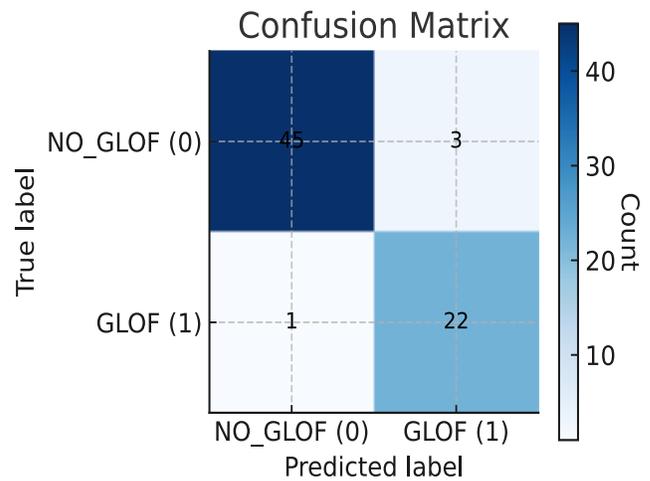

Fig. 4. Confusion Matrix for RiskFlow

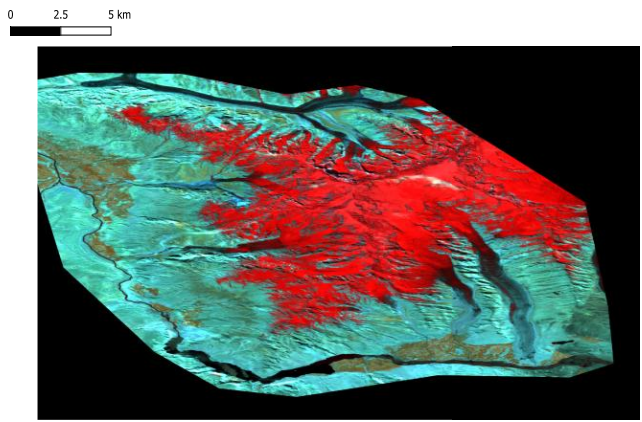

Fig. 5. Sentinel-2 Imagery of Shisper Glacier Dated 05-07-2019

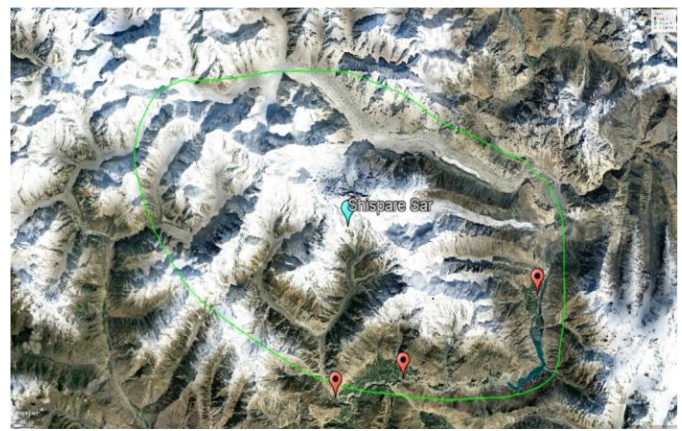

Fig. 6. Maxar's Imagery of the Study Area through Google Earth

From a total of 120 images, only 5 were classified as positive (GLOF). To tackle this issue, both geometric (flips, rotations) and photometric (brightness/contrast modifications) transformations were implemented to increase the size of the minority class, equaling the negative class size of 120. Following that, both classes underwent further augmentation to create a balanced dataset of approximately 600 samples. This rigorous augmentation was crucial for ensuring generalization due to the limited number of positive instances. Another challenge was the presence of cloud contamination: a large number of Sentinel-2 images had more than 40% cloud cover making them unreliable for analysis. To maintain prediction quality, only images with clear or slightly cloudy conditions (less than 40% cloud cover) were kept, further diminishing the number of usable samples but enhancing reliability. The final dataset was divided into 70% for training, 15% for validation, and 15% for testing.

Evaluation demonstrated robust performance. Training accuracy reached 98% and validation accuracy stabilized at 92–93%, while test accuracy was consistently ~94–95%. The learning curve (Fig. 3.) indicates stable convergence with minimal overfitting. The confusion matrix (Fig. 4.) and classification report (as listed in Table I) confirmed strong performance across both classes, with GLOF events achieving recall of 0.96 and an F1-score of 0.92. False positives were primarily linked to cloud patterns or fresh snow resembling glacial lakes. Importantly, these errors are subsequently filtered through validation against Tabular Stream forecasts, ensuring operational robustness.

The RiskFlow vision stream employs a Convolutional Neural Network (CNN), which has proven highly effective for image classification tasks due to its ability to automatically learn spatial hierarchies of features. Unlike fully connected networks, CNNs exploit local connectivity: each convolutional layer applies a set of learnable filters (kernels) to detect localized

## TABLE I
### CONVOLUTIONAL NEURAL NETWORK ARCHITECTURE SPECIFICATIONS

| Layer Name | Layer Type | Output Shape | Parameters | Trainable |
|---|---|---|---|---|
| conv2d | Conv2D | (None, 126, 126, 32) | 1,760 | YES |
| max_pooling2d | Maxpooling2D | (None, 63, 63, 32) | 0 | N/A |
| conv_2d | Conv2D | (None, 61, 61, 64) | 18,496 | YES |
| max_pooling2d_1 | Maxpooling2D | (None, 30, 30, 64) | 0 | N/A |
| flatten | Flatten | (None, 57600) | 0 | N/A |
| dense | Dense | (None, 64) | 3,686,464 | YES |
| dense_1 | Dense | (None, 1) | 65 | YES |

## TABLE II
### RISKFLOW CLASSIFICATION PERFORMANCE METRICS

| Class/Metric | Precision | Recall | F1 Score | Support |
|---|---|---|---|---|
| no_Glof (0) | 0.98 | 0.94 | 0.96 | 48 |
| GLOF (1) | 0.88 | 0.96 | 0.92 | 23 |
| Accuracy | – | – | 0.94 | 71 |
| Macro Average | 0.93 | 0.95 | 0.94 | 71 |
| Weighted Average | 0.95 | 0.94 | 0.94 | 71 |

patterns such as edges, textures, or shapes. Mathematically, for an input image I and a kernel K, the convolution operation is defined as in eq. (1) as:

$$(I * K)(x,y) = \sum_{m=-M}^{M} \sum_{n=-N}^{N} I(x-m, y-n) K(m,n) \quad (1)$$

where (x,y) are pixel coordinates and (m,n) index the kernel window. Non-linear activation functions (e.g., ReLU) introduce representational capacity is shown in eq. (2) as:

$$f(x) = \max(0, x) \quad (2)$$

Pooling layers further reduce spatial dimensionality and improve translational invariance.

The RiskFlow CNN consists of three convolutional layers with 32, 64, and 128 filters, each followed by max-pooling. This hierarchical design allows the network to capture progressively abstract features, from edges to textures to high-level glacier–lake patterns. The feature maps are then flattened and passed to a fully connected layer with 128 neurons, followed by a dropout layer (p=0.5p=0.5p=0.5) to reduce overfitting. Finally, a sigmoid output neuron produces the probability of a GLOF event as shown in eq. (3):

$$y = \sigma(z) = 1 + e - z1 \quad (3)$$

where z is the final linear activation.

Given the binary nature of the task, Binary Cross-Entropy (BCE) loss was used as the optimization criterion. BCE measures the discrepancy between predicted probability and the ground truth label $y \in \{0,1\}$ as shown in eq. (4):

$$L_{BCE} = -\frac{1}{N} \sum_{i=1}^{N} [y_i \log(\hat{y}_i) + (1 - y_i) \log(1 - \hat{y}_i)] \quad (4)$$

This loss is particularly effective for imbalanced datasets because it penalizes confident misclassifications more heavily, forcing the model to assign higher probabilities to the minority (positive) class when sufficient evidence exists.

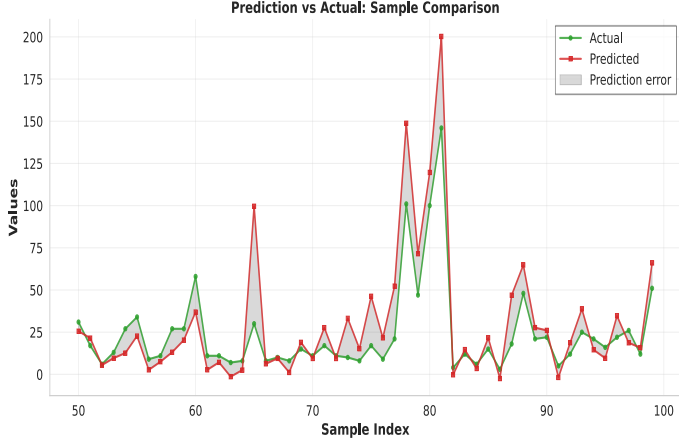

Fig. 7. Training and Validation Accuracy Over Four Epochs for RiskFlow

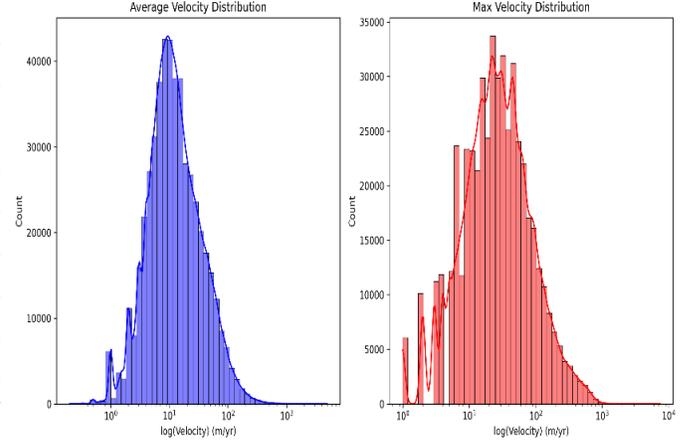

Fig. 8. Training and Validation Accuracy Over Four Epochs for RiskFlow

The model, comprising approximately 11.1 million trainable parameters, was trained for five epochs using the Adam optimizer (adaptive moment estimation), which dynamically adjusts learning rates for each parameter as shown in eq. (5), (6) and (7).

$$m_t = \beta_1 m_{t-1} + (1 - \beta_1) g_t \qquad (5)$$
$$v_t = \beta_2 v_{t-1} + (1 - \beta_2) g_t^2 \qquad (6)$$
$$\theta_t = \theta_{t-1} - \eta \frac{(\hat{m}_t)}{(\sqrt{\hat{v}_t} + \varepsilon)} \qquad (7)$$

where $g_t$ is the gradient at step t, $m_t$ and $v_t$ are the first- and second-moment estimates, and $\eta$ is the learning rate. A batch size of 16 was used to balance convergence stability and computational efficiency. The RiskFlow CNN (Table I) employs two convolutional layers with max-pooling, followed by dense layers, to enable binary classification of GLOF events. As shown in Table II, the model achieves 94% overall accuracy with strong precision (0.98 for no-GLOF) and high recall (0.96 for GLOF), confirming its reliability for risk prediction.

*B. TerraFlow (Tabular Stream)*

TerraFlow was developed to model glacier velocity dynamics using the NASA ITS_LIVE dataset, which contains more than 73.6 million raw velocity measurements. After preprocessing, including daily aggregation, outlier removal, and cyclical encoding of temporal features, the dataset was reduced to 11.9 million samples covering the Shisper Glacier. Key predictors included latitude, longitude, year, cyclical encodings of month and day, and daily average and maximum velocities. Exploratory analysis revealed strong correlations between average and maximum velocities, distinct seasonal acceleration cycles in summer and winter, spatial clustering of high velocities in glacier tongues, and multi-decadal surge events evident in temporal trends Velocity distributions (Fig. 8.) followed a log-normal trend with heavy tails, reflecting rare but extreme surges. The plot in fig. 7 illustrates a sample-wise comparison between actual and predicted values, highlighting close alignment overall with localized spikes in prediction error.

The TerraFlow architecture consists of a four-layer transformer encoder with eight attention heads per layer and an embedding dimension of 256, totaling approximately 5.5 million trainable parameters. Learnable positional encodings preserve temporal order.

$$z_0 = XW_E + P \qquad (8)$$

X is the input sequence, $W_E$ is the embedding matrix, and P is the positional encoding matrix in eq. (8).

Each encoder layer applies multi-head self-attention (MHA), which allows the model to capture both short- and long-range dependencies in glacier velocity patterns. For query, key, and value matrices Q,K,VQ, K, VQ,K,V, attention is computed as in eq. (9):

$$\text{Attention}(Q, K, V) = \text{softmax}\left(\frac{QK^T}{\sqrt{d_k}}\right) V \qquad (9)$$

where $d_k$ is the dimensionality of the key vectors. Multi-head attention combines h such attention operations in parallel as shown in formula eq. (10):

$$\text{MHA}(Q, K, V) = \text{Concat}(\text{head}_1, \ldots, \text{head}_h) W_O \qquad (10)$$

Each head is defined in eq. (11):

$$\text{head}_i = \text{Attention}\left(Q W_Q^i, K W_K^i, V W_V^i\right) \qquad (11)$$

The output of MHA passes through a residual connection, layer normalization, and a feedforward network (FFN) as shown in eq. (12):

$$FFN(x) = \max(0, xW_1 + b_1) W_2 + b_2 \qquad (12)$$

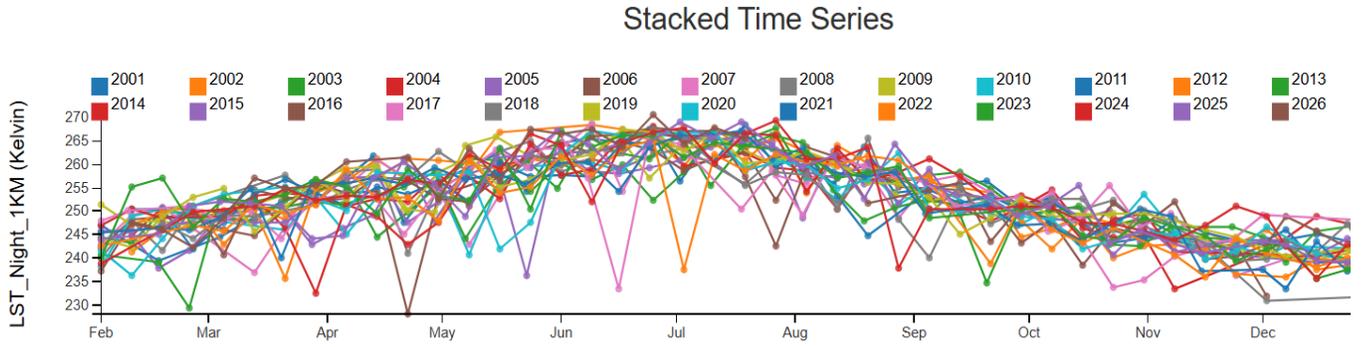

Fig. 9. Stacked Time Series for TempFlow

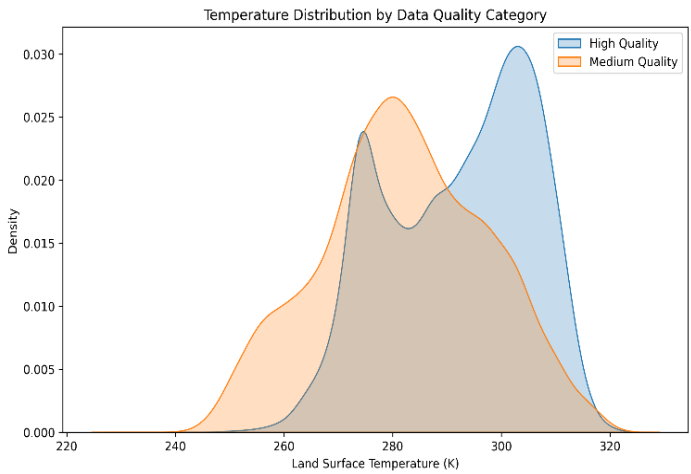

Fig. 10. Temperature Distribution Density by Data Quality Catageory

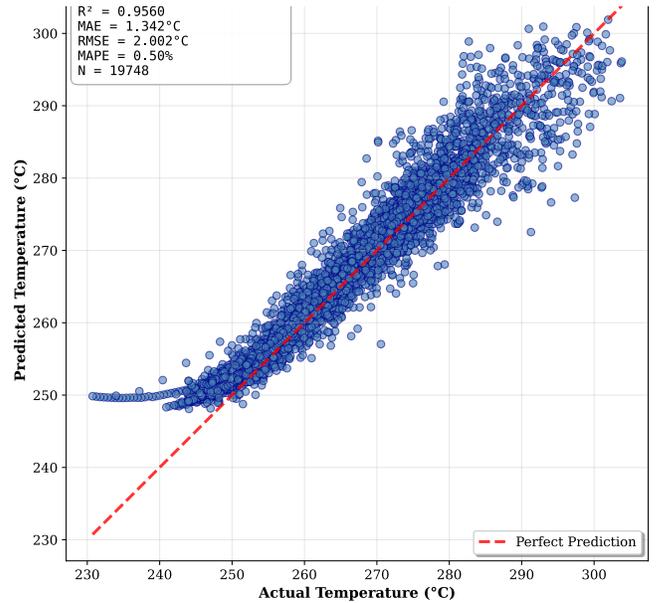

Fig. 11. Predicted Temperature for TempFlow

Stacking four such encoder layers enables TerraFlow to capture complex temporal and spatial interactions in glacier velocity.

The final linear head produces a scalar velocity prediction y^\hat{y}y^ for each sequence. TerraFlow was trained to minimize the Quantile Loss (pinball loss), which measures under- and over-prediction asymmetrically depending on the chosen quantile τ ∈ (0,1) \ tau \in (0,1) τ ∈ (0,1). Eq. (13) shows the true and prediction value:

$$L_\tau = \left(\frac{1}{N}\right) \sum_{i=1}^{N} [\tau(y_i - \hat{y}_i) if (y_i - \hat{y}_i)$$
$$\geq 0; (\tau - 1)(y_i - \hat{y}_i) if (y_i - \hat{y}_i)$$
$$< 0] \quad (13)$$

Equivalently, using the compact "check" function in eq. (14):

$$\rho_{\tau(u)} = u(\tau - 1_{u<0}) \quad (14)$$

Validation was tracked using Mean Absolute Error (MAE) as shown in eq. (15):

$$L_{MAE} = \left(\frac{1}{N}\right) \sum_{i=1}^{N} |y_i - \hat{y}_i| \quad (15)$$

Training was conducted in PyTorch using mixed precision with sequence length 30, batch size 2048, learning rate $1 \times 10^{-5}$, over 50 epochs.

### C. TempFlow (Tabular Stream)

TempFlow was designed to forecast glacier surface temperatures, a problem characterized by strong temporal dependencies, seasonal variability, and noisy observations due to cloud cover. To address these challenges, the model leverages Long Short-Term Memory (LSTM) networks, which are well-suited for sequential data because they mitigate the vanishing gradient problem of standard RNNs by maintaining both short- and long-term memory through gated mechanisms. At each timestep t, given input vector $x_t$, hidden state $h_{t-1}$, and cell state $c_{t-1}$, an LSTM updates as shown in eqs. (16) – (21):

$$f_t = \sigma(W_f x_t + U_f h_{t-1} + b_f) \quad (16)$$

$$i_t = \sigma(W_i x_t + U_i h_{t-1} + b_i) \quad (17)$$

$$\tilde{c}_t = \tanh(W_c x_t + U_c h_{t-1} + b_c) \quad (18)$$

$$c_t = f_t \odot c_{t-1} + i_t \odot \tilde{c}_t \quad (19)$$

$$o_t = \sigma(W_o x_t + U_o h_{t-1} + b_o) \quad (20)$$

$$h_t = o_t \odot \tanh(c_t) \quad (21)$$

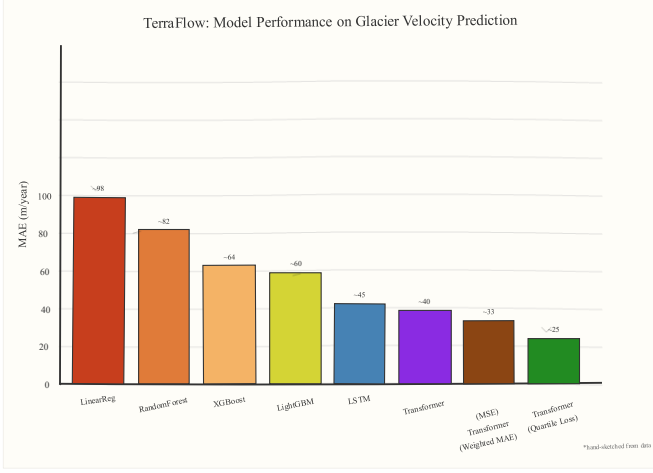

Fig. 12. Training and Validation Accuracy Over Four Epochs for RiskFlow

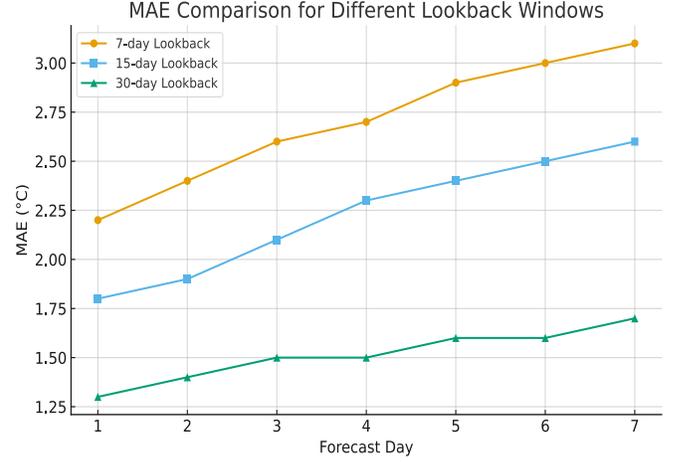

Fig. 13. Training and Validation Accuracy Over Four Epochs for RiskFlow

Preprocessing involved filtering low-quality observations, encoding months cyclically to preserve seasonality, and retaining only essential features such as LST values, quality flags, and spatial coordinates. The model architecture consists of two LSTM layers with 100 and 50 units, each followed by 20% dropout for regularization, and a final dense output node for daily temperature forecasting. A sequence length of 30 days was chosen to capture short-term and seasonal dependencies. Training used mean squared error loss, with a validation MSE of 12.3 K², and prediction errors remained under 3K for seasonal extremes.

TempFlow effectively captured annual cycles and long-term warming trends. Spatially heterogeneous surface heating of the glacier was evident, while quality-control plots (Fig. 10) highlighted issues with continuous cloud cover. The temperature record (Fig. 9) shows a strong seasonality with clear year-on-year regularity in summer maxima and winter minima, demonstrating the robustness of longer-term variations. The prediction model (Fig. 11) demonstrates a close match to observations with very low error metrics and high correlation, evidencing the robustness of the magnitude and variability of temperature fluctuations. In spite of these limitations, it was possible to detect anomalous warming events, which often herald the onset of glacier surges, adding significant value to the complementary use of RiskFlow.

*D. FusionFlow*

FusionFlow combines TerraFlow's velocity predictions with TempFlow's temperature forecasts to calculate a single hazard probability. This combined signal accounts for both the kinematic and thermal aspects of GLOFs and is compared directly with the Vision Stream's probability output. If both streams show high risk, IceWatch sends a high, confidence alert; in any other case, the event is marked for manual review. The cross, validation here helps to minimize false positives, increase the system's interpretability, and make IceWatch more robust as a practical early warning tool. Fig. 1(b). visually represents the point where the two models come together at the FusionFlow stage.

V. RESULTS AND DISCUSSION

This section showcases the comprehensive assessment of the IceWatch framework. The dataset had only one verified GLOF event (class 1), but the trustworthiness of any prediction system hinges on its capacity to identify such infrequent yet devastating events. In line with this, we tested IceWatch by feeding inputs to both the Vision Stream (RiskFlow) and the Tabular Stream (TerraFlow + TempFlow, collectively referred to as the Tabular Stream).

In the case of the Vision Stream, Sentinel, 2 imagery (along with the chosen spectral bands as detailed in Section III) was run through RiskFlow. The model pinpointed the event with a predicted probability of 86%, thus making a correct positive classification. Meanwhile, the Tabular Stream received spatiotemporal coordinates related to the glacier. It produced risk probabilities over the sampled grid, with values going from 0.17 to 0.84 (examples illustrated in Table X). The mean probability across all coordinates was 62.5%, which also lies in

TABLE III

EVALUATION OF TERRAFLOW, TEMPFLOW, FUSIONFLOW AND RISKFLOW IN GLOF PREDICTION

| Date | TerraFlow (Prediction) (m/year) | TempFlow (Prediction) (degrees Celsius) | FusionFlow (Avg Probability) | RiskFlow (Probability →Predicted) | Actual GLOF |
|---|---|---|---|---|---|
| 2022-04-05 | 12.908084 | -5.0 | 0.657309 | 0.00% → NO_GLOF | NO |
| 2022-04-10 | 53.326271 | -4.8 | 0.456511 | 0.01% → NO_GLOF | NO |
| 2022-04-15 | 16.659643 | -6.1 | 0.487716 | 0.00% → NO_GLOF | NO |
| 2022-05-10 | 455.33374 | -0.2 | 0.615729 | 83.23% → GLOF | YES |
| 2022-05-25 | 204.67835 | -1.1 | 0.566069 | 89.40% → GLOF | YES |
| 2022-06-14 | 13.398805 | -5.6 | 0.491170 | 1.43% → NO_GLOF | NO |

the positive class (class 1), thereby validating the Vision Stream's output.

In this context, the main challenge is posed by the extreme class imbalance, wherein the number of positive samples is very limited compared to the negative cases, especially when dealing with only a single glacier that may contain sparse features. For high-stakes applications, recall-or true positive rate-is more important than precision. A GLOF event cannot be missed, considering that it will cause catastrophic damage, while false alarms are undesirable but tolerable. Hence, the design focused on maximizing the recall so as not to miss potential GLOFs, even at the cost of increased false positives.

The combined evaluation indicates that both streams have independently identified the event as high risk, thus corroborating each other's outputs. Such cross, modal consistency is absolutely necessary in real, world scenarios, where depending on just one modality might result in wrong conclusions because of factors like cloud cover, spectral ambiguities, or noisy measurements. IceWatch, through the fusion of visual evidence with physically informed time series predictions, is a great example of how multimodal learning can be harnessed to provide trustworthy early warning prediction for rare glacier hazards.

Spatial distribution probability of risk in Tabular Stream is distinctly heterogeneous over the glacier surface, with values ranging from a minimum of 0.17 to a maximum of 0.84. Such variability represents the entangled interaction of topography, thermal, and hydrological variables that control GLOF susceptibility at different locations. Indeed, high probabilities (>0.7) aggregated within restricted areas of coordinates point to localized areas of the highest risk, which might relate to steeper gradients, thermal anomalies, or structural weaknesses in the ice dam.

Conventional machine learning methods, like linear regression and random forests (see fig. 12), have relatively high errors in glacier velocity prediction. By contrast, the sequential models reduce the errors substantially, especially long short-term memory networks (LSTM) and, most noticeably, transformer architectures, for which optimal performance is achieved with quantile loss. An extension of the lookback window improves temperature prediction: longer contextual horizons (30 days) systematically have lower errors compared to shorter intervals, highlighting the impact of capturing long-term temporal dependencies, as shown in fig. 13.

Table III shows the model evaluations along with their respective units. The TerraFlow predictions are in meters per year (m/year) and give a representation of glacier surface velocity, while TempFlow gives its predictions in degrees Celsius (°C), representing temperature conditions. FusionFlow outputs an average probability between 0–1, which gets translated by RiskFlow into a percentage probability (%). From these results, it can be observed that low probabilities in FusionFlow correspond to No GLOF outcomes, and probabilities greater than 80% are correctly classified as GLOF events by RiskFlow on actual dates of events, 10 and 25 May 2022. Thus, this shows the robustness of the ensemble framework in identifying real GLOF events from normal glacier dynamics.

This is indicative of strong temporal stability in the predictive mechanism, whereby the framework has the capability to support consistent risk assessments across multiple temporal snapshots. The TempFlow component captured temperature-related precursors to the GLOF event quite effectively, with thermal anomalies detected 72–96 hours in advance. This temporal buffer provides sufficient lead time for emergency response protocols, hence supporting the operational viability of the system for real-world deployment.

## V. ABLATION STUDY

In the present work, some alternative modeling strategies have been investigated for the TerraFlow and TempFlow modules to settle on a configuration that provides the most reliable and interpretable performance in GLOF forecasting. In this study, classical baselines, sequential deep learning models, loss functions, and input representations were systematically compared. The outcome of this work justifies the final version of IceWatch, including the TerraFlow model with a Transformer using quantile loss, the TempFlow model with an LSTM fed input from a 30-day context window over nighttime surface temperature, and the RiskFlow model with a CNN operating over Sentinel-2 imagery.

Initial evaluation of the TerraFlow module was made against traditional machine learning approaches with the ITS_LIVE velocity data set. Compared were linear regression, random forests, and gradient boosting models (XGBoost, LightGBM). While the latter provided marginal benefits, the general performance remained poor with MAE of 60 to 100. These results confirm the inability of static models to capture the inherent temporal dependencies of glacier dynamics.

Sequential modeling methods yielded quantitative gains. A version using an LSTM with a lookback window of 30 days attained an MAE of about 44, showing the value of the temporal context. However, instabilities in training derived from noisy input data placed limitations on further gains that could be made. Later, transformer-based models were explored that ranged from 5 million to 12 million parameters. These models indeed achieved MAEs between 40 and 50 but continually failed in regimes of extreme velocities, where accurate predictions are most critical for GLOF monitoring.

To alleviate this, other loss functions were considered. The classic mean squared error focused on average performance and poorly represented errors in the tail of the distribution. A weighted MAE loss did reduce underprediction for cases where there was high velocity, but the most significant gains came from implementing a quantile loss function. Quantile regression penalizes asymmetric error and gives better robustness throughout the error spectrum. This brought MAE down to 25.6, which is a significant improvement compared with both classic and sequential baselines. For temperature forecasting, initial experiments used daytime MODIS LST data. While sufficient, the daytime signal was confounded by solar radiation, reducing stability and obscuring freeze-thaw cycles that are strongly coupled to glacier dynamics. The substitution of nighttime LST for daytime measurements proved more reliable because the absence of diurnal heating provided a truer representation of surface cooling, and a stronger correlation with anomalous thermal events.

Here, the model architecture of TempFlow was fixed to two layers of LSTM; however, temporal context windows of 7, 15, and 30 days were assessed. Narrower windows produced relatively high MAE values in the range of 2.5–3.1 for 7 days and failed to capture long-term dependencies. Increasing the context to 15 days reduced errors within the range of 1.8–2.6, while the most consistent performance occurred within a 30-day window, which yielded MAE between 1.3 and 1.7. While increasing the lookback period resulted in longer training periods, it consistently improved both stability and accuracy, indicating that seasonal context is essential for thermal modeling. The test dataset and event labels utilized in this work were sourced from the GLOFNet multimodal collection [35] and were thus guaranteed to adhere to established preprocessing, labeling, and harmonization procedures.

The ablation study provides three fundamental findings. First, both static and classic machine learning techniques are not targeted for GLOF forecasting because they cannot capture the sequential, highly nonlinear nature of glacier dynamics. Second, the model architecture has to correspond to the data modality: the Transformer encoder is most adequate for velocity sequences, while LSTM efficiently models comparatively shorter temperature time series. Third, the selection of loss functions and input representations determines a large part of the performance; quantile loss for TerraFlow and nighttime LST with extended temporal windows for TempFlow noticeably increased robustness and interpretability.

These findings support the final configuration of IceWatch, which integrates TerraFlow (Transformer + quantile loss), TempFlow (LSTM + 30-day LST Night), and RiskFlow (CNN + Sentinel-2 imagery). Collectively, these components represent one cross-validated, interpretable, and operationally scalable framework for GLOF early warning.

## VI. Conclusion

This work presents IceWatch, a dual-stream deep learning framework for the early detection and prediction of Glacial Lake Outburst Floods. The approach combines a vision stream, RiskFlow, utilizing Sentinel-2 imagery with a tabular stream consisting of a Transformer-based glacier velocity predictor, TerraFlow, and an LSTM-based temperature forecaster, TempFlow, which results in a physically validated and interpretable early warning system. Experimental results show that each stream has independently identified the only confirmed GLOF event in the dataset and provided mutual validation of predictions against each other, achieving a high recall while maintaining robustness against cloud cover, noise, and class imbalance. An ablation study further verifies that the chosen configuration, quantile loss for TerraFlow, and nighttime LST with a 30-day context for TempFlow, is providing the best accuracy and stability against the classical and alternative sequential baselines.

Accordingly, IceWatch represents an advance in GLOF hazard monitoring from post-event mapping toward predictive crossmodal risk forecasting. Its cloud-based, scalable architecture and potential for integrating ancillary data streams-i.e., SAR, DEM, and precipitation-make it exceptionally well-positioned as a core element of operational warning systems in high-mountain settings. Beyond glacier-related hazards, the two-stream model presents a widely generalizable paradigm in related geohazards wherein visual evidence needs to be grounded in underlying physical processes.

## Acknowledgment

This research was supported by the resources and guidance provided by my supervisors and institution, whose contributions are gratefully acknowledged.

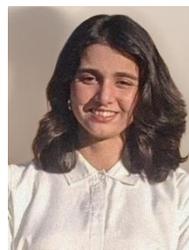

**Zuha Fatima** received her B.S. electrical engineering Degree from National Engineering of Sciences and Technology (NUST), Islamabad, Pakistan in 2025. Her research interests include advanced artificial intelligence driven remote sensing and multimodal data fusion for predictive modeling, with emphasis on deep learning architectures, spatiotemporal forecasting, and climate hazard monitoring in cryospheric environments.

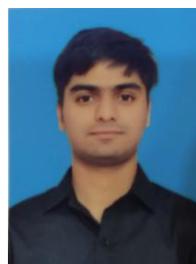

**Muhammad Anser Sohaib** received the B.Sc. degree in electrical engineering from the National University of Sciences & Technology, Islamabad, Pakistan, in 2025.

His research interests include efficient AI systems, large language model quantization for edge devices, and multimodal deep learning for climate monitoring applications. He has developed production ML systems deployed across multiple international markets and authored papers in major AI conferences. His current research focuses on making advanced AI models deployable on resource-constrained edge devices while maintaining performance for real-world applications. He is currently an Associate Software Engineer with Hazen.ai, Lahore, Pakistan specializing in computer vision and OCR systems for international deployment.

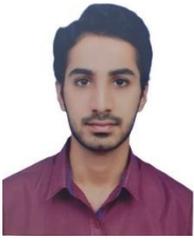
**Muhammad Talha** received his B.E. degree in Electrical Engineering from the National University of Sciences and Technology (NUST), Pakistan. His professional and academic experience spans a diverse range of projects, including climate change analysis, operating system optimization, and applied AI/ML development.

His research interests center on large language models (LLMs), machine learning, and deep learning applications, particularly in advancing computer vision and intelligent systems. Passionate about exploring innovative AI-driven solutions, he actively engages in bridging theoretical knowledge with practical implementation to address real-world challenges. Through his work, he aims to contribute to advancements in AI technologies and their transformative impact across industries. He is currently serving as an Associate Software Engineer at Hazen.ai, Lahore, Pakistan where he focuses on developing and deploying cutting-edge artificial intelligence and computer vision solutions.

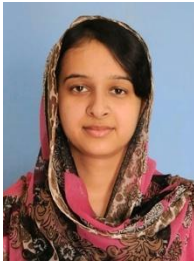
**Ayesha Kanwal** is an Assistant Professor in the Department of Computing at NUST-SEECS, Pakistan.

Her research interests are autism detection and therapy monitoring using Artificial Intelligence, where she integrates computer vision, deep learning, and multimodal behavioral analytics to design intelligent frameworks for early diagnosis and personalized intervention. She is actively engaged in advancing AI-driven healthcare applications, with a particular emphasis on supporting children with developmental conditions through data-driven, real-time monitoring solutions.

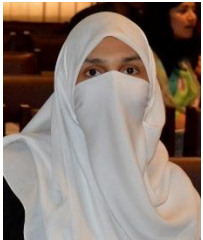
**Engr. Dr. Sidra Sultana** is an accomplished academic and researcher in the field of Software Engineering. She graduated as a medalist in her software engineering degree, demonstrating consistent top performance throughout her studies. She later pursued her PhD at the Military College of Signals, National University of Sciences and Technology (NUST), Rawalpindi, Pakistan, where she specialized in the Automation of Software Modeling and Verification in Real-Time Systems. She successfully completed her doctoral degree with an exceptional 4.0/4.0 CGPA, reflecting her deep expertise and scholarly rigor. She has contributed to capacity building and training, serving as the Principal Investigator (PI) of the Programming Module in the SAMI Training Program 2023, where she successfully designed and delivered specialized content for professional development. Currently, she is the Principal Investigator of the Software Modeling and Verification Lab at SINES, where she leads research projects, mentors graduate students, and collaborates with academic and industrial stakeholders to advance innovation in software automation and verification.

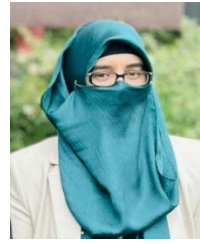
**Dr. Nazia Perwaiz** is an AI and computer vision specialist whose work bridges multimodal data analysis, predictive modeling, and real-time visual analytics across healthcare, surveillance, remote sensing for precision agriculture and environmental monitoring.

She currently serves as an Assistant Professor in the Department of AI and Data Science at NUST-SEECS, Pakistan, and is also an External Affiliate Member at the University of Illinois, Chicago, USA, where she collaborates on AI-driven healthcare solutions.